\def\year{2022}\relax
\documentclass[letterpaper]{article} % DO NOT CHANGE THIS
\usepackage{aaai22}  % DO NOT CHANGE THIS
\usepackage{times}  % DO NOT CHANGE THIS
\usepackage{helvet}  % DO NOT CHANGE THIS
\usepackage{courier}  % DO NOT CHANGE THIS
\usepackage[hyphens]{url}  % DO NOT CHANGE THIS
\usepackage{graphicx} % DO NOT CHANGE THIS
\usepackage{natbib}  % DO NOT CHANGE THIS AND DO NOT ADD ANY OPTIONS TO IT
\usepackage{caption} % DO NOT CHANGE THIS AND DO NOT ADD ANY OPTIONS TO IT
\DeclareCaptionStyle{ruled}{labelfont=normalfont,labelsep=colon,strut=off} % DO NOT CHANGE THIS
\usepackage{booktabs} 
\usepackage{multirow}
\usepackage{graphicx}
\usepackage{arydshln}
\usepackage[ruled,vlined,noend]{algorithm2e}

\usepackage[]{todonotes}

\title{Zero-shot Commonsense Question Answering with Cloze Translation and Consistency Optimization}

\author{Zi-Yi Dou, Nanyun Peng}

\affiliations{
    University of California, Los Angeles \\
    { \{zdou,violetpeng\}@cs.ucla.edu }
}
\usepackage{bibentry}
\begin{document}

\maketitle

\begin{abstract}
Commonsense question answering (CQA) aims to test if models can answer questions regarding commonsense knowledge that everyone knows. Prior works that incorporate external knowledge bases have shown promising results, but knowledge bases are expensive to construct and are often limited to a fixed set of relations. In this paper, we instead focus on better utilizing the \textit{implicit knowledge} stored in pre-trained language models. 
While researchers have found that the knowledge embedded in pre-trained language models can be extracted by having them fill in the blanks of carefully designed prompts for relation extraction and text classification, it remains unclear if we can adopt this paradigm in CQA where the inputs and outputs take much more flexible forms. 
To this end, we investigate four translation methods that can translate natural questions into cloze-style sentences to better solicit commonsense knowledge from language models, including a syntactic-based model, an unsupervised neural model, and two supervised neural models. In addition, to combine the different translation methods, we propose to encourage consistency among model predictions on different translated questions with unlabeled data. 
We demonstrate the effectiveness of our methods on three CQA datasets in zero-shot settings. We show that our methods are complementary to a knowledge base improved model, and combining them can lead to state-of-the-art zero-shot performance. Analyses also reveal distinct characteristics of the different cloze translation methods and provide insights on why combining them can lead to great improvements.\footnote{Code/dataset is available at \url{https://github.com/PlusLabNLP/zero_shot_cqa}.}
\end{abstract}

\section{Introduction}
Commonsense knowledge consists of widely known facts that humans use to reason and react to everyday situations. Recently, empowering machines with such commonsense reasoning abilities has become an active research topic~\cite{lin2019kagnet,bosselut2019comet,lv2020graph} and various commonsense question answering (CQA) benchmarks have been constructed~\cite{zellers2018swag,sap2019atomic,zellers2019recognition}. Different from other types of QA tasks, CQA usually does not require domain-specific knowledge and sophisticated natural language understanding. Rather, it relies on inference over implicit commonsense knowledge that is not given in the QA contexts.

\begin{figure}[t]
\centering
\includegraphics[width=0.5\textwidth]{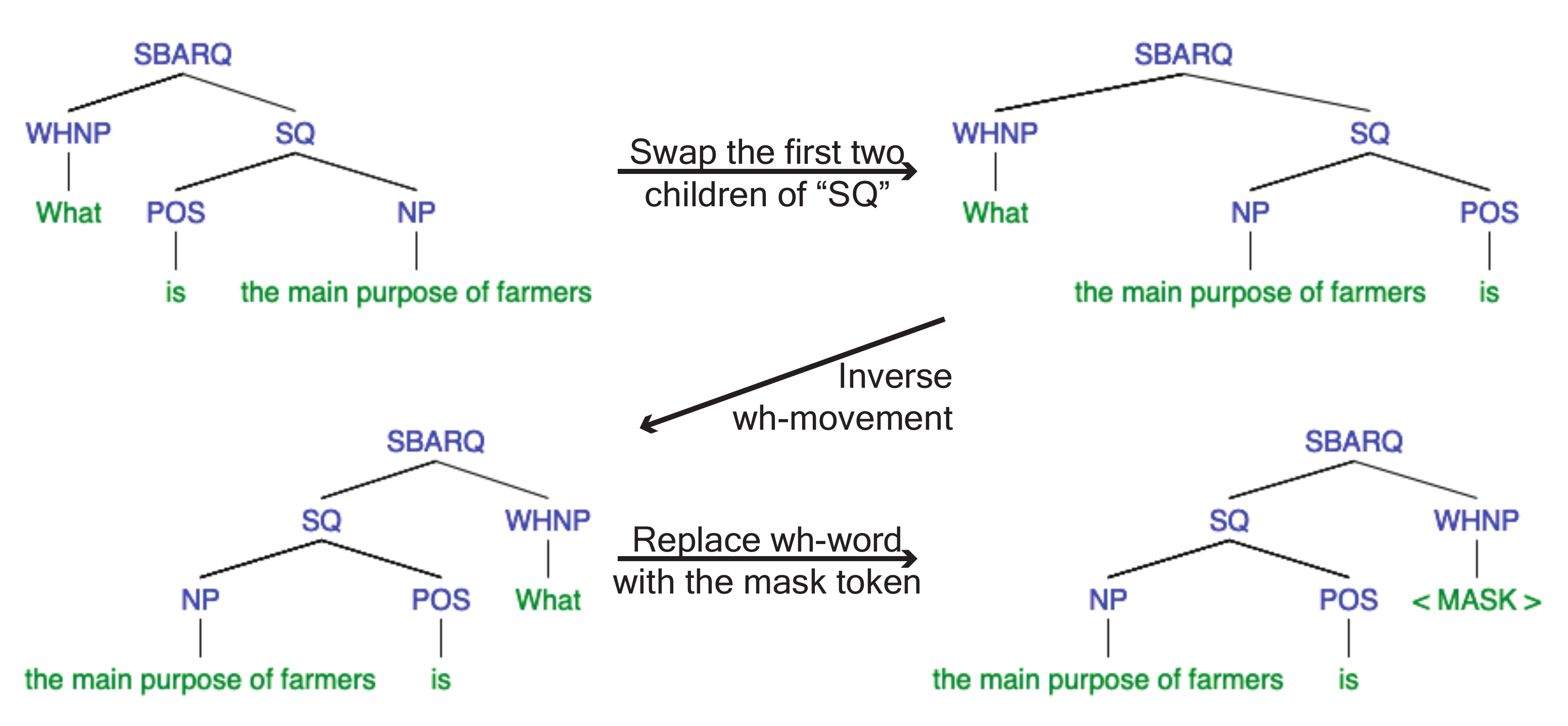}
\caption{An example of natural-to-cloze translation with our syntactic-based method. `SQ' is defined as the subconstituent of questions excluding wh-word or wh-phrase.}
\label{fig:rule}
%\vspace{-1.5em}
\end{figure} 

To tackle this problem, researchers have attempted to construct commonsense knowledge bases~\cite{vrandevcic2014wikidata,speer2017conceptnet,sap2019atomic}, which can be integrated into downstream models~\cite{bosselut2019dynamic}. However, knowledge bases are often limited to a pre-defined set of relations and are expensive to construct. On the other hand, language models (LMs; e.g.~\citet{devlin2019bert,liu2019roberta,lan2019albert}) pre-trained on large textual corpora are easy to extend to more data and allow users to query about an open class of relations. In addition, it has been demonstrated that pre-trained LMs contain a certain amount of world knowledge implicitly~\cite{roberts2020much,talmor2020olmpics} which can be extracted by having LMs fill in the blanks of carefully designed  prompts~\cite{petroni2019language,zhou2020evaluating,jiang2020can}. However, these previous work only focuses on the settings where there is a fixed set of relations or output classes, thus knowledge can be induced by designing a limited amount of hand-crafted or automatically-generated rules. For example, to obtain a birthplace of one person X, we can just have an LM fill in the blank of `X was born in \_'. 
In contrast, natural questions are much more flexible and it is non-trivial to design general rules to transform different natural questions into cloze forms. 
How to better solicit implicit knowledge in the pre-trained LMs
for CQA is an open question and no previous work has explored cloze translation for CQA to our knowledge. %while LMs have shown great performance in different CQA tasks~\cite{}, it remains unclear that if the LMs are just modeling annotator bias of individual datasets during the task-specific fine-tuning stage instead of learning how to utilize the implicitly learned knowledge effectively~\cite{miral,kaixin}.

In this paper, we propose to better exploit the knowledge embedded in LMs for CQA by translating natural commonsense questions into ``fill-in-the-blank'' cloze sentences (see Figure~\ref{fig:rule} for an example). We investigate four translation methods, including 1) a syntactic-based model that performs a sequence of syntactic transformations on the source questions; 2) an unsupervised neural sequence-to-sequence (seq2seq) model that does not require any natural-cloze question pairs inspired by~\citet{lewis2019unsupervised}; 3) a supervised seq2seq model~\cite{lewis2020bart} that is trained on our constructed dataset of natural-cloze question pairs; 4) a sequence tagging model~\cite{omelianchuk-etal-2020-gector} that performs operations such as word insertions and deletions on the source natural questions and transforms them into the target cloze questions. In addition, to combine the strengths of different translation models, we propose a consistency optimization objective which encourages the consistency between model predictions on the different translated cloze questions of the same instance using only unlabeled data.

We mainly focus on the zero and few-shot settings as commonsense QA should be questions that any human can answer without specific training, so we want to equip models with similar ability. Moreover, these settings are robust measures of the models' general reasoning abilities~\cite{ma2020knowledge}. 
We experiment on three CQA datasets, including CommonsenseQA~\cite{talmor2019commonsenseqa}, OpenbookQA~\cite{Mihaylov2018CanAS}, and SocialIQA~\cite{sap2019socialiqa}. Results demonstrate that our cloze translation methods can achieve significant improvements on both zero and few-shot settings, and the consistency optimization objective can lead to further improvements. In addition, we show that our methods are complementary to a state-of-the-art knowledge base improved method and can bring extra gains. 
Analyses provide insights on distinct characteristics of the different cloze translation methods and why combining them can lead to greater improvements. Finally, we demonstrate that our methods can be beneficial in high-resource settings when the models are trained with both natural questions and our translated cloze-form queries.

\section{Methods}

We first present four different cloze translation methods, discuss how we make use of the cloze questions, then illustrate how we combine them using consistency optimization on unlabeled data.\footnote{We focus on multiple-choice commonsense question answering. Formally, given a natural question $q$ and a set candidate answers $\{a_i\}$, the model needs to select the most probable answer. }

\subsection{Cloze Translation}
We investigate four methods for cloze translation:
\paragraph{Syntactic-based Rewriting.} Transforming natural questions to cloze questions can be understood as a series of syntactic transformation rules. While it can be nontrivial to design a perfect set of rules~\cite{heilman2010good}, here we adopt some simple heuristics and our general idea is shown in Figure~\ref{fig:rule}. We use the constituency parser in~\cite{Joshi2018ExtendingAP} to get the part-of-speech tags and syntactic structure of the input questions. The syntactic-based rewriting model does not require any training data, but it can be inflexible as it is hard to take all kinds of natural questions into consideration.

\begin{algorithm}[t]
\SetAlgoLined
\SetKwProg{Fn}{Function}{}{end}
\Fn{Transform(root)}{
 \If{$root$ has no children}{
    \Return {$root$[`sentence']}
 }
 %sentence = `' \\
 \For{$child$ in $root$[`children']}{
    $next\_child$ = $child.right\_sibling$ \\
    \If{$next\_child$[`nodeType'] is `SQ'}
    {Do inverse wh-movement on $child$ and replace its `wh'-word with `\texttt{[MASK]}'}
    \ElseIf{$child$[`nodeType'] is `SQ'}
    {Swap\_first\_two\_children($child$)}
    \Else{Transform($child$)}
 }
 \Return {$root$[`sentence']}
}
 \caption{\label{alg:ref1}Our syntactic-based rewriting method (`SQ' is defined as the subconstituent of questions excluding wh-word or wh-phrase)}
\end{algorithm}

Specifically, we mainly consider two cases in this paper. First, if there exist nodes with the type `SQ' in the input sentence, where `SQ' is defined as the constituent of questions excluding wh-word or wh-phrase, we apply Algorithm~\ref{alg:ref1} on the sentence. To illustrate, Algorithm~\ref{alg:ref1} mainly swaps the first two children of the `SQ' node, then performs an inverse wh-movement on the inputs, and finally replaces the wh-word with the mask tooken. Note that when doing the wh-word replacement, we replace `what', `who', `which' with `\texttt{[MASK]}'; `why' with `because \texttt{[MASK]}'; `how' with `by \texttt{[MASK]}'; `where' with `at \texttt{[MASK]}'; `when' with `when \texttt{[MASK]}'. Otherwise, if there is no `SQ' node in the tree, we search through the sentence and replace the first wh-word with `\texttt{[MASK]}'.

  \begin{table*}[th]
  \centering
  \resizebox{\textwidth}{!}{
  \small
  \begin{tabular}{@{}p{8cm}@{\ \ }p{8cm}@{}}
  \toprule
  \bf Source & \bf Target \\
  \midrule
 What do people aim to do at work?  & People aim to \texttt{[MASK]} at work.\\ What could go on top of wood? & \texttt{[MASK]} could go on top of wood.\\
 Where could you find a toilet that only friends can use? & You could find a toilet that only friends can use at \texttt{[MASK]}. \\
 How is riding a bike getting it to move? & Riding a bike is getting it to move by \texttt{[MASK]}. \\
 Why would you be watching TV instead of doing something else? & You would be watching TV instead of doing something else because of \texttt{[MASK]}. \\
 \bottomrule
    \end{tabular}
    }
    %\vspace{-2mm}
    \caption{ \label{tab:data-example} Samples from the created cloze translation data.}
    %\vspace{-1.5em}
  \end{table*}
\paragraph{Unsupervised Seq2Seq.}
\citet{lewis2019unsupervised} have shown that we can perform unsupervised cloze translation by training neural seq2seq models with denoising auto-encoding and iterative back-translation objectives on unparallel natural and cloze question data. Their unsupervised cloze translation method~\cite{lewis2019unsupervised} borrows some ideas from unsupervised neural machine translation methods~\cite{lample2018unsupervised,lample2018phrase}. Concretely, first, they create a cloze question corpus by masking noun phrases and named entities in statements sampled from Wikipedia, and a natural question corpus by mining questions containing some common wh-words from CommonCrawl. Then, they train both cloze-to-natural and natural-to-cloze translation methods with denoising auto-encoding and iterative back-translation objectives as in unsupervised machine translation. The denoising auto-encoding objective first masks some of the tokens in the source questions, and the model is trained to reconstruct the original questions. For the iterative back-translation objective, a target-to-source model is first used to translate a target question into the source side, then a source-to-target model is trained to output the original target question given the translated source sentence, and this process will be repeated in both directions iteratively. Their model architecture uses a 4-layer Transformer~\cite{vaswani2017attention} encoder and a 4-layer Transformer decoder.

While~\citet{lewis2019unsupervised} mainly focus on cloze-to-natural translation and using it to perform data augmentation for question answering, a by-product of their method is a natural-to-cloze translation model, and here we directly use their pre-trained model.\footnote{\url{https://dl.fbaipublicfiles.com/UnsupervisedQA/sentence_ne.tar.gz}} The unsupervised model is much more flexible than the syntactic-based rewriting one, but the translated questions can be deviated from the original inputs due to the lack of supervision signals and the uncontrollable nature of seq2seq models.

\paragraph{Supervised Seq2Seq.} To provide models with supervisions, we manually create a dataset of natural-cloze question pairs. Concretely,
we manually translate all the natural questions in the original CommonsenseQA training and development data into cloze questions. We create a (8,500/1,221/1,241) split as in our main experiments. It takes a person around 40 hours to construct such a dataset. We sample several examples from the dataset as shown in Table~\ref{tab:data-example}. We can see that there exist different kinds of transformation rules and previous methods on designing the prompts for pre-trained language models cannot be applied in commonsense question answering.

We fine-tune BART-Large~\cite{lewis2020bart}, a representative seq2seq model, on the dataset and perform natural-to-cloze translation. The inputs and outputs are the natural and cloze questions respectively. The model is trained with maximum likelihood estimation objective and we employ beam search during decoding. We choose BART-Large as our supervised seq2seq model because it is widely used in text generation tasks such as text summarization. BART is based on the Transformer model~\cite{vaswani2017attention} and is pre-trained by corrupting documents and then optimizing a reconstruction loss. Its architecture consists of 12 Transformer encoding and decoding layers.

\paragraph{Supervised Sequence Tagging.} While seq2seq models have been the de facto choice for many sequence generation tasks, cloze translation mainly involves word movements, deletions, and insertions which do not require a whole re-writing. Therefore, we also formulate cloze translation as a sequence tagging problem. A tagging model identifies which
words need to be changed
and modifies them with pre-defined word-level transformations (\textit{e.g.} keep, delete, append), which may generate more faithful cloze questions than seq2seq models. The sequence tagging task has been widely investigated in the task of grammatical error correction and here we train GECToR~\cite{omelianchuk-etal-2020-gector}, a popular model in grammatical error correction, on our constructed dataset.

The GECToR model~\cite{omelianchuk-etal-2020-gector} employs a Transformer encoder and its parameters are initialized with RoBERTa. They have pre-defined token-level transformations. For example, given a source and target sentence `A ten years old boy go school' and `A ten-year-old boy goes to school.', it first pre-processes the pair to convert it to a sequence of transformation rules. Concretely, first, they map each token from the source sentence to subsequence of tokens from the target sentence: [A $\rightarrow$ A], [ten $\rightarrow$ ten, -], [years $\rightarrow$ year, -], [old $\rightarrow$ old], [go $\rightarrow$ goes, to], [school $\rightarrow$ school.]. Then, they will find token-level transformations that convert the source tokens to the target tokens and there is only one transformation for each source token: [A $\rightarrow$ KEEP], [ten $\rightarrow$ MERGE\_HYPHEN], [years $\rightarrow$ NOUN\_NUMBER\_SINGULAR], [old $\rightarrow$ KEEP], [go $\rightarrow$ VERB\_FORM\_VB\_VBZ], [school $\rightarrow$ APPEND\_DOT]. Because there is only a single tag for each token, this method is not suitable for all the situations. To solve the problem, they propose to process the pairs iteratively and at each step there is one single tag for each token.

We can see that after the pre-processing, sequence generation is turned to a sequence classification task. Therefore, we can encode the entire input sentence using its encoder and feed the encoded representations to a classifier. The classifier will decide which transformation rule to apply for each token. We refer readers to their codebase\footnote{\url{https://github.com/grammarly/gector}} for more details.

\subsection{Answer Prediction} %Predicting on the Cloze Questions}
\label{sec:2.1}
Once we have the cloze question $x$ for a natural question $q$, we can replace the mask token in $x$ with each of the candidate answers in $\{a_i\}$, and feed each replaced sentence $r_i= \langle r_{i1}, \cdots, r_{in} \rangle $ to a pre-trained LM. The pre-trained LM can assign a score for $r_i$ with
\begin{equation}
    s_{i} = \frac{1}{n}\sum_{k=1}^n \log p (r_{ik} | r_{i}  ),
\end{equation}
and $r_i$ with the highest score is treated as our prediction.\footnote{We find that averaging the output logits is sometimes better than averaging the log probabilities, and in the experiments we select the best strategy based on the development set.} We can apply \textit{softmax} on the resulting scores $\{s_{i} \}$ to get the prediction probabilities.

\subsection{Consistency Optimization}
The above methods can generate different cloze questions $\{x_j\}$ for $q$. Instead of picking a single one for LMs, we propose a consistency optimization objective to combine them with unlabeled data.

\label{sec:2.2}
Inspired by work on multi-view learning~\cite{wang21naacl}, we encourage the consistency between predictions on different cloze translations for the same question. For each training instance, we first ensemble the model predictions: formally, given each cloze question $x_j$ and its candidate answers $\{a_i\}$, we obtain the score $\{s_{ij} \}$ for $\{a_i\}$ as in the previous section, then apply \textit{softmax} on $\{s_{ij} \}$ to get the prediction probabilities $\{p_{ij}\}$; then, we sum the probabilities for each answer from different translation as $p_i=\sum_j p_{ij}$, and take the answer with the highest probability as the ensemble prediction $a^*$.

We then treat $a^*$ as the pseudo-label to supervise the LM in a self-training manner. Concretely, we fine-tune the LM to maximize the probability of $x_j$ with its mask token replaced with $a^*$ while minimizing the probability of all the other candidate answers in $\{a_i\}$ with the cross-entropy loss.

Note that the consistency optimization objective does not need any gold labels, making it possible to be integrated into the zero-shot and few-shot settings.

  \begin{table*}[th]
  \centering
%  \resizebox{\textwidth}{!}{
  \small
  \begin{tabular}{@{}l@{\ \ }p{7.5cm}@{\ \ }p{7cm}@{}}
  \toprule
  \bf Method & \bf Natural Question & \bf Cloze Question \\
  \midrule
  Syntactic-based & Where is a good idea but not required to have a fire extinguisher? & But is a good idea not required to have a fire extinguisher at \texttt{[MASK]}. \\
  Unsup. Seq2Seq &What island country is ferret popular? & The island country is a popular ferret in \texttt{[MASK]}. \\
  Sup. Seq2Seq & Blue read material outside of his comfort zone because he wanted to gain what? & James read material outside of his comfort zone because he wanted to gain \texttt{[MASK]}.\\
  Sup. Tag &  Where is a human likely to go as a result of being hungry? & A is likely to go to \texttt{[MASK]} as a result of being hungry.\\
 \bottomrule
    \end{tabular}
%    }
    %\vspace{-2mm}
    \caption{ \label{tab:example} Example failure cases of our translation methods sampled from the CommonsenseQA dev set.}
    %\vspace{-1.5em}
  \end{table*}
  %\vspace{-2mm}
\section{Experiment}
%\vspace{-2mm}
%\subsection{Settings}
We experiment on CommonsenseQA~\cite{talmor2019commonsenseqa}, OpenbookQA~\cite{Mihaylov2018CanAS}, and SocialIQA~\cite{sap2019socialiqa}. Because the standard test set labels of CommonsenseQA are unavailable, we create a (8,500/1,221/1,241) split for (train/dev/test) following~\citet{lin2019kagnet,wang2020connecting}. We use the standard splits for the other datasets. We compare with two methods that do not use any knowledge base, including 1) natural questions (`Base'), which are directly concatenated with each answer choice and then fed to pre-trained LMs; 2) self-talk~\cite{shwartz2020unsupervised}, which gets additional background for commonsense questions by querying LMs. We also report the results of~\citet{ma2020knowledge}, which is a state-of-the-art zero-shot CQA model using knowledge bases to construct CQA datasets automatically. We use ALBERT-xxlarge-v2~\cite{lan2019albert} as the base LM. We will illustrate the details of our experimental setup in the following paragraphs.

 \begin{table*}[t]
     \small
  \centering
  \begin{tabular}{@{}lcccccccc@{}}
  \toprule
  \multirow{2}{*}{\bf Method} &\multicolumn{2}{c}{\bf CommonsenseQA} & &\multicolumn{2}{c}{\bf OpenbookQA} & &\multicolumn{2}{c}{\bf SocialIQA}  \\
   \cmidrule{2-3} \cmidrule{5-6}  \cmidrule{8-9}
    & \bf dev & \bf test  &&  \bf dev & \bf test & & \bf dev & \bf test\\
   % \midrule
  %\multicolumn{4}{l}{\textit{\textbf{Zero-Shot}} } \\
   \midrule
  \multicolumn{4}{l}{\textit{\textbf{Methods without Knowledge Base}} } \\
  \midrule
  \multicolumn{4}{l}{\textit{{Baseline}} } \\
  \midrule
  Base (ALBERT)  & 31.14 & 28.52 &&  31.80 & 33.00 && 41.71 & 40.47  \\
  \hdashline
  self-talk (GPT2) & 31.53 & 29.74 && 28.40 & 30.80 && 45.34 & 44.47\\
  self-talk (ALBERT) & 15.89 & 17.49 && 22.20 & 19.40 && 26.25 & 26.48\\
  \midrule
  \multicolumn{4}{l}{\textit{{Ours (ALBERT-based)}} } \\
  \midrule
  Syntactic-based rewriting & 50.94 & 48.67 && 41.60 & 39.80 && 44.11 & 42.00 \\
  Unsup. Seq2Seq & 43.49 & 42.86 && 40.00 & 39.20 && 40.94 & 38.80 \\
  Sup. Seq2Seq & 51.60 & 49.00 && 39.00 & 39.80 && 44.73 & 41.41 \\
  Sup. Tag & 50.86 & 48.51 && 39.00 & 38.60 && 41.53 & 40.78 \\
   \hdashline
   Ensemble & 54.62 & 51.57 && 41.00 & 39.20 && 44.11 & 42.04\\
   Consistency* & 64.07 $\pm$ 0.14 & 61.08 $\pm$ 0.35 && 50.27 $\pm$ 0.57 & 49.87 $\pm$ 0.90 && 54.13 $\pm$ 0.99 & 54.21 $\pm$ 1.37\\
 % \midrule
 % \multicolumn{4}{l}{\textit{{Combining Cloze-Transformed Data}} } \\
 % \midrule
 % Ensemble & 54.62 & 51.57 && 41.00 & 39.20 && 44.11 & 42.04\\
 % Consistency & 64.13 & 61.24 && 50.20 & 49.80 && 55.42 & 56.07\\ 
 % GPT2 & 40.37 & 37.23 && 31.20 & 29.40 && 40.79 & 38.08 \\
 %  Bert & 28.91 & 25.87 \\
 % Roberta & 31.45 & 28.61 \\
 % GPT2-XL & 41.03 & 35.61 \\
 % T5-Large & 31.94 & 31.02 \\
 \midrule
  \multicolumn{4}{l}{\textit{\textbf{Methods Using Knowledge Base}} } \\
  \midrule
  \multicolumn{4}{l}{{\textit{Baseline}} } \\
  \midrule
  \citet{ma2020knowledge} (RoBERTa) & 68.63 & 66.88 && 34.80 & 38.00 && 56.04 & 51.93 \\
  \citet{ma2020knowledge} (ALBERT) & 66.50 & 64.87 && 45.40 & 48.00 && 51.02 & 52.28 \\
\midrule
  \multicolumn{4}{l}{{\textit{Ours (ALBERT-based)}} } \\
  \midrule
 \citet{ma2020knowledge} + Consistency*   & \bf 69.73 $\pm$ 0.16 & \bf 67.38 $\pm$ 0.44 && \bf 58.27 $\pm$ 0.25 & \bf 54.27 $\pm$ 0.41 && \bf 59.85 $\pm$ 0.72 & \bf 59.88 $\pm$ 0.97\\ 
  \bottomrule
    \end{tabular}
    \caption{ \label{tab:zero-shot} Accuracy (\%) in zero-shot settings. `*' indicates that we run the experiments three times with different random seeds. For self-talk and~\citet{ma2020knowledge}, we try both ALBERT and the best LMs used in their papers. The best scores are in \textbf{bold}.}
    %\vspace{-1.5em}
  \end{table*}

  \subsection{Experimental Setup}

\paragraph{Datasets.} For the CommonsenseQA dataset~\cite{talmor2019commonsenseqa}, because its test set is not publicly available, the predictions for it can only be evaluated once every two weeks via the official leaderboard. Therefore, following previous work~\cite{lin2019kagnet,wang2020connecting}, we separate the training data into training and test sets consisting of 8,500 and 1,241 instances respectively. We use the standard development set consisting of 1,221 instances. The OpenbookQA~\cite{Mihaylov2018CanAS} dataset consists of 5,957 multiple-choice questions with 4,957 training, 500 development, 500 testing instances. While it provides a small ``book'' of 1,326 core science facts, we do not include this additional information because our focus is on the implicitly learned knowledge in pre-trained language models. The SocialIQA~\cite{sap2019socialiqa} dataset contains 33,410 training, 1,954 development, 2,224 testing instances, the aim of which is to probe the emotional and social intelligence of models in a variety of everyday situations.

\paragraph{Cloze Translation Methods.} For the unsupervised cloze translation method, we use the pre-trained model (sentence cloze boundaries, named entity answers) provided by~\citet{lewis2019unsupervised}. For the seq2seq model, we follow the setting in text summarization on XSUM and fine-tune the BART-Large model on the training set of our cloze data for 15k steps with a batch size of 16,384 tokens and a learning rate of 3e-5. For the sequence tagging model, we fine-tune the RoBERTa-based GECToR model on our cloze translation data with default parameters.\footnote{https://github.com/grammarly/gector} We select the models that achieve the best BLEU scores on the development set for cloze translation.

\paragraph{Consistency Optimization.} For the consistency optimization objective, we use the training data of each dataset without using their labels. We encourage the model prediction consistency on the data generated by syntactic-based rewriting, supervised seq2seq, and supervised sequence tagging model. The models are trained with a learning rate of 1e-5 for 2k/1k/2k steps for CommonsenseQA/OpenbookQA/SocialIQA respectively. 

\paragraph{Zero-shot Settings.} In the zero-shot settings, for the baseline ALBERT-xxlarge-v2 model, we directly concatenate the questions and answers together, and feed the concatenated sentences to the model to get the language modeling scores. For the self-talk baseline, we try both GPT2-Large and ALBERT-xxlarge-v2 for querying the external contexts and getting the language modeling scores using the default parameters. For~\citet{ma2020knowledge}, we use their constructed CWWV data that utilizes three knowledge bases: ConceptNet, WordNet, and Wikidata, then we train both ALBERT-xxlarge-v2 and RoBERTa-Large on the CWWV data with their default parameters. 

\paragraph{Few-shot Settings.} For the few-shot settings, we randomly sample 16/32/64/128 datapoints from the training data and fine-tune ALBERT-xxlarge-v2 on both the natural and cloze translated data. Similar to the consistency optimization objective, a question is concatenated with each answer choice and then fed to the model to get its corresponding score. The scores for all the choices are then normalized using \textit{softmax} to get the prediction probability for the choices, then we use the cross-entropy loss on the prediction probability to train the model. The learning rate is set to 1e-5 and the number of epoch is set to 10 or 20, selected on the development sets. We run the experiments three times with different random seeds for each setting.

%\paragraph{High-resource Settings.} In the high-resource settings, we fine-tune ALBERT-xxlarge-v2 on both the natural and cloze translated data with all the training data. The models are trained with a learning rate of 1e-5 for 2k/1k/2k steps for CommonsenseQA/OpenbookQA/SocialIQA. For ensemble, we apply \textit{softmax} on the prediction scores for each model, and add the prediction probabilities together. We try to ensemble 3 models trained on natural data, 3 models trained on cloze data, and 2 models trained on natural and cloze data respectively.
  
\subsection{Main Results}
In this section, we will present the main results of our methods in both zero- and few-shot settings. 

\subsubsection{Methods without Knowledge Base}

We first compare with methods that utilize knowledge base. Table~\ref{tab:zero-shot} shows that cloze translation can generally improve the zero-shot performance of ALBERT significantly across settings (one exception is unsupervised seq2seq on SocialIQA), demonstrating that the knowledge in LMs can indeed be better extracted with cloze questions. Unsupervised seq2seq is the least effective translation method, potentially due to its lack of supervisions. 
Our models cannot outperform self-talk on SocialIQA, possibly because self-talk manually designs task-specific question transformation rules for SocialIQA, which inserts strong supervision into their models.

We then combine all the translation methods except unsupervised seq2seq. We find that directly ensembling them as in Section~\ref{sec:2.2} cannot always lead to improvements, but consistency optimization can improve the performance by a large margin across settings. This demonstrates that it is highly nontrivial to combine the strengths of different translation methods and our designed objective is one very effective way of combining them. Note that after using the consistency optimization objective, our method can even achieve comparable performance with \citet{ma2020knowledge} that leverages external knowledge base on two datasets.

\paragraph{Methods Using Knowledge Base}
We also try to compare our models with a method that uses external knowledge base and achieves state-of-the-art performance in zero-shot settings. While our methods cannot always outperform~\citet{ma2020knowledge}, which is intuitive considering that our model is given less information, we can combine~\citet{ma2020knowledge} with our method. Specifically, because~\citet{ma2020knowledge} mainly use knowledge base to construct datasets, we can first fine-tune LMs on their constructed data and then use our cloze translation and consistency optimization methods on the commonsense question answering datasets. As we can see from Table~\ref{tab:zero-shot}, this combination strategy can lead to the best performance across datasets, indicating the complementarity of the two methods.

\begin{figure}[t]
\centering
\includegraphics[width=0.5\textwidth]{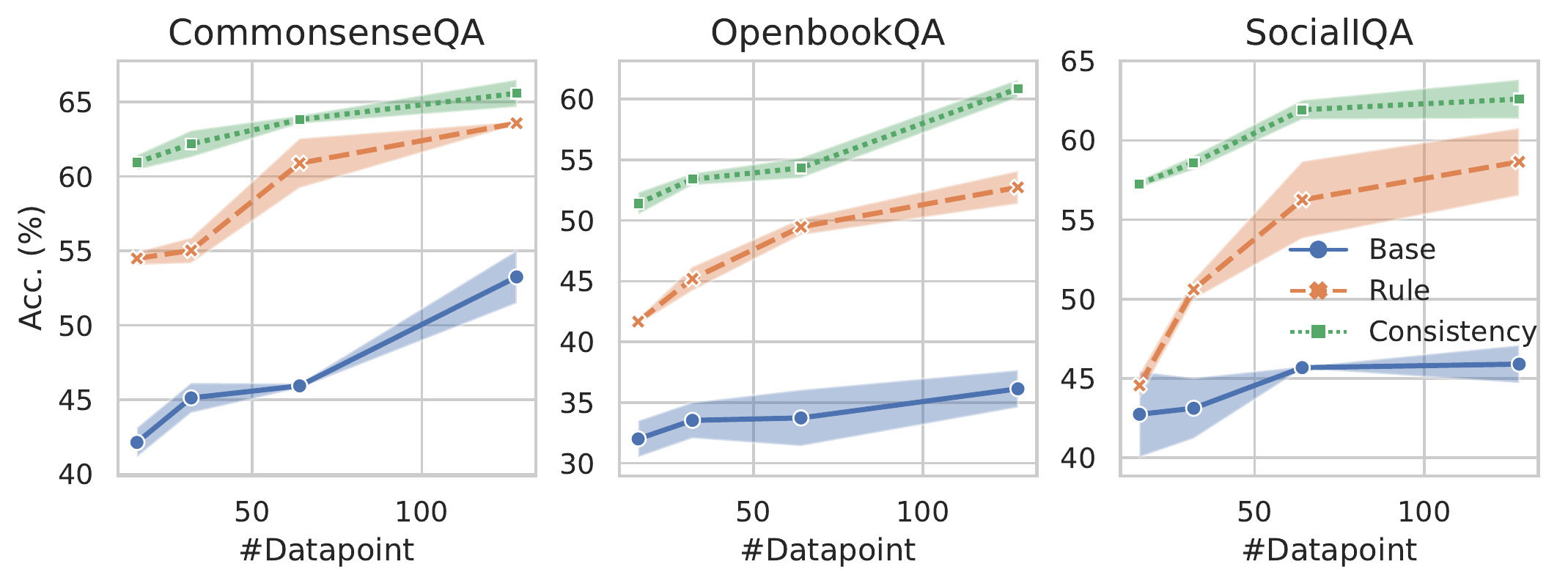}
%\vspace{-2em}
\caption{Accuracy (\%) in few-shot settings. We run the experiments three times for each setting.}
\label{fig:few}
%\vspace{-1em}
\end{figure}

\paragraph{Few-shot Evaluations.} We also experiment in few-shot settings where only 16/32/64/128 instances are available. In this part, we mainly compare the baseline with the syntactic-based translation because 1) it does not need any supervisions; 2) it works well in zero-shot settings as shown in Table~\ref{tab:zero-shot}. As illustrated in Figure~\ref{fig:few}, our syntactic-based translation method consistently outperforms the baseline which is trained on natural questions and consistency training can also be helpful in these settings. Also, it is interesting to note that zero-shot performance of cloze translation is better than supervised models trained with 100 natural questions. 

\subsection{Analysis}
We then perform several analyses on our methods.
%[{'1': 86, 'wr': 6, 'un': 8}, {'1': 87, 'un': 1, 'faith': 11, 'wr': 1}, {'1': 86, 'un': 6, 'faith': 6, 'wr': 2}, {'1': 52, 'faith': 7, 'wr': 32, 'un': 9}]
\begin{table}[t]
  \centering
  \small
  \resizebox{\columnwidth}{!}{
  \begin{tabular}{@{}r@{\ \ }r@{\ \ }r@{\ \ }r@{\ \ }r@{}}
  \toprule
  \bf Method & \bf Correct & \bf Unnatural & \bf Unfaithful & \bf Wrong \texttt{[MASK]} \\
  \midrule
  Syntactic-based & 86 &  8 & 0 & 6 \\
  Unsup. Seq2Seq & 52 & 9 & 7 & 32 \\
  Sup. Seq2Seq & 87 & 1 & 11 & 1 \\
  Sup. Tag & 86 &  6 & 6 & 2 \\
 \bottomrule
    \end{tabular}
    }
    %\vspace{-2mm}
    \caption{ \label{tab:error} Error counts of our methods. `Wrong \texttt{[MASK]}' means the position of \texttt{[MASK]} is wrong. }
    %\vspace{-2em}
  \end{table}

\paragraph{Translation Errors.} We first try to analyze the errors of different cloze translation methods. To this end, we randomly sample 100 questions from the CommonsenseQA dataset and perform human evaluation on the translation errors as in Table~\ref{tab:example} and~\ref{tab:error}. As we can see from the tables, the unsupervised seq2seq method is the least effective one as it can often generate meaningless questions, which can explain why it performs in worst in the main experiment section (Table~\ref{tab:zero-shot}. The syntactic-based method, on the other hand, is inflexible, thus it can generate unnatural sentences or put \texttt{[MASK]} at the wrong place when dealing with complex syntactic structures. In the example in Table~\ref{tab:example}, we can see that it can generate the sentence ``the island country is a popular ferret in [MASK]'' which is hard to parse and quite unnatural.

The supervised methods are more much flexible. However, the supervised seq2seq method can sometimes generate unfaithful outputs. Interestingly, in Table~\ref{tab:example}, it replaces the person name `Blue' in the original question with `James', possibly because `James' appears more frequently in the training data. Note that even though the output is unfaithful, it does not affect the correct answer choice. For the supervised tagging model, because it mainly deals with word deletions and insertions, sometimes over-deletions or insertions may occur, resulting in unfaithful or unnatural sentences. But it should be noted that the supervised tagging method can generate more faithful outputs than the seq2seq method, confirming our previous hypothesis.

We can see that different cloze translation methods have rather distinct characteristics, indicating that the translation outputs can be rather diverse, which can be the reason why our consistency optimization objective can greatly improve the model performance across settings as in Table~\ref{tab:zero-shot}.

%UQA: What island country is ferret popular?
%['The island country is a popular ferret in [MASK] .']
%BART: Blue read material outside of his comfort zone because he wanted to gain what?
%'James read material outside of his comfort zone because he wanted to gain [MASK].'
%rule: Where is a good idea but not required to have a fire extinguisher?
%'but is a good idea not required to have a fire extinguisher at [MASK] .
%GECTOR: Where is a human likely to go as a result of being hungry?
%A is likely to go to [MASK] as a result of being hungry
%{'what': 68, 'where': 28, 'who': 2, 'why': 1, 'which': 1}

\paragraph{High-resource Settings.} We also test the model performance on natural and cloze questions in high-resource settings where all the labeled training data are used. In the high-resource settings, we fine-tune ALBERT-xxlarge-v2 on both the natural and cloze translated data with all the training data. The models are trained with a learning rate of 1e-5 for 2k/1k/2k steps for CommonsenseQA/OpenbookQA/SocialIQA. For the ensemble, we apply \textit{softmax} on the prediction scores for each model, and add the prediction probabilities together. We try to ensemble 3 models trained on natural data, 3 models trained on cloze data, and 2 models trained on natural and cloze data respectively.

\begin{figure}[t]
\centering
\includegraphics[width=0.5\textwidth]{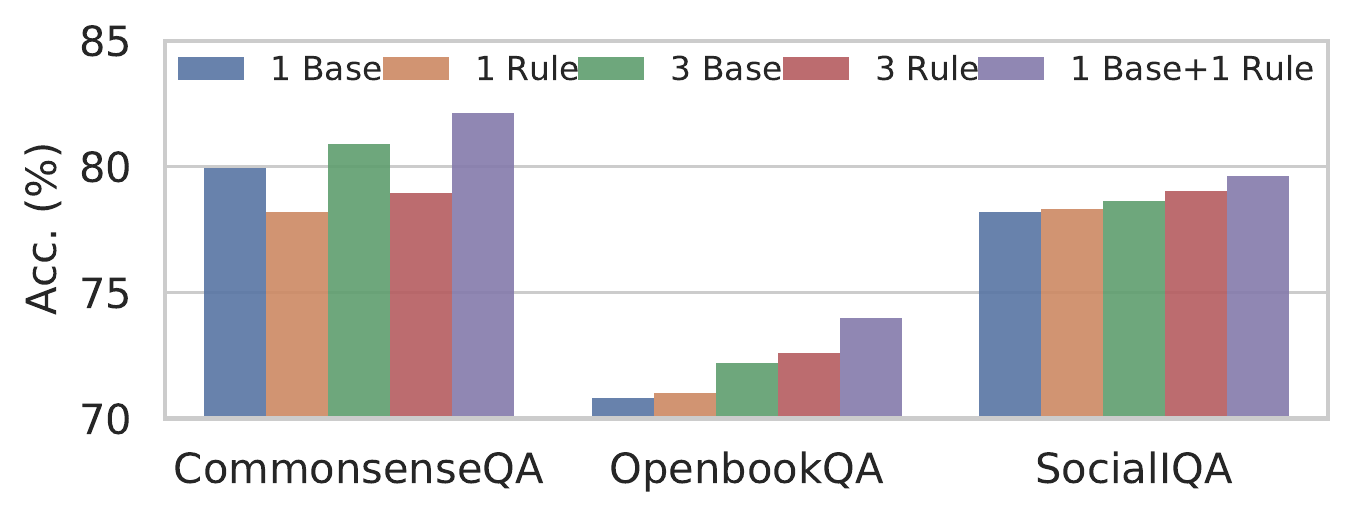}
\caption{Accuracy (\%) on natural (`Base') and rule-translated (`Rule') data in high-resource settings. Ensembling each of the models trained on different data (`1 Base + 1 Rule') is better than ensembling 3 models trained on the same data (`3 Base' and `3 Rule').}
\label{fig:high}
%\vspace{-1em}
\end{figure}

As in Figure~\ref{fig:high}, our model (`1 Rule') cannot always outperform the baseline (`1 Base'). We hypothesize that this is due to the translation errors as we have analyzed in the previous part. Concretely, the translation errors can alter the meaning of the original questions, and training on these noisy data can lead to degraded performance. However,  because of the diversity among different translation methods, we can ensemble each of the models trained on different data (`1 Base + 1 Rule'), which is better than ensembling 3 models trained on the same data with different random seeds (`3 Base' and `3 Rule').

\begin{figure}[t]
\centering
\includegraphics[width=0.5\textwidth]{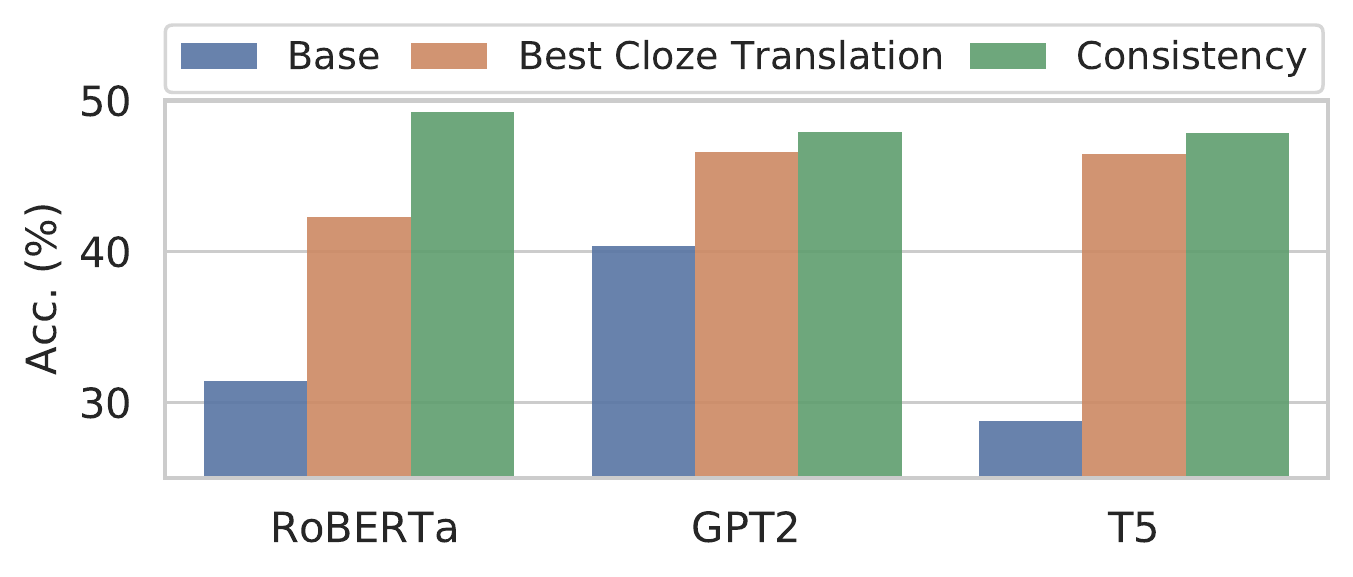}
 %\vspace{-2mm}
\caption{Performance of RoBERTa-Large, GPT2-Large, T5-Large on the CommonsenseQA dev set.}
 %\vspace{-1em}
\label{fig:lm}
\end{figure} 

\paragraph{Applicability to Other LMs.} All of the previous experimental results are obtained using the ALBERT model. In this part, we also test if our methods are applicable to other LMs as well. As shown in Figure~\ref{fig:lm}, our methods can improve all the pre-trained LMs. Also, we can see that our model has a tendency to favor bidirectional LMs (\textit{e.g.} ALBERT and RoBERTa) than uni-directional models.  One possible reason is that when using cloze translation, the LM probability of all the words will be affected in bidirectional LMs, while only the probability of words appearing after \texttt{[MASK]} are changed for unidirectional LMs. For example, cloze translation improves GPT2 the least possibly because it is unidirectional and only the probability of words succeeding \texttt{[MASK]} are affected. Similarly, consistency training improves T5 marginally because it is a sequence-to-sequence model and the target side are the label words. Therefore, only the label word probabilities will be affected which can limit the model performance. %More results are in Appendix D.%, with a tendency to favor bidirectional LMs such as ALBERT and RoBERTa. More results are in Appendix.

\section{Related Work}
%Finally, we overview two lines of related work: %, including commonsense question answering and knowledge extraction from pre-trained language models.
 %Researchers have created different benchmarks~\cite{zellers2018swag,zhou2019going,sakaguchi2020winogrande,sap2019atomic,zellers2019recognition,talmor2019commonsenseqa,lin2020commongen,bisk2020piqa,sap2019socialiqa}, 
\paragraph{Commonsense Question Answering.} Researchers have created different benchmarks~\cite{zellers2018swag,zhou2019going,sakaguchi2020winogrande,sap2019atomic,zellers2019recognition,talmor2019commonsenseqa,lin2020commongen,bisk2020piqa,sap2019socialiqa}, which motivates the research on commonsense question answering. Most previous work on commonsense question answering tries to incorporate knowledge base during training ~\cite{bosselut2019dynamic,bosselut2019comet,ye2019align,ma2020knowledge} or during inference~\cite{bauer2018commonsense,lin2019kagnet,xu2020fusing,lv2020graph,wang2020connecting}. For example, \citet{ma2020knowledge} use knowledge bases to automatically construct data for commonsense question answering and train the models on their constructed datasets. \citet{xu2020fusing} try to fuse knowledge into models by having the models being able to attend to knowledge bases such as ConceptNet. Different from these methods, we focus on better utilizing the knowledge embedded in pre-trained LMs. 

Recently,~\citet{shwartz2020unsupervised} propose a self-talk framework that can induce commonsense knowledge from LMs by iteratively querying them to discover additional background knowledge given a question. They also manually design dataset-specific rules for cloze translation. In this paper, we treat it as our baseline and take a step further, developing more principled ways of cloze translation and achieving better performance than their methods.

\paragraph{Knowledge Exploitation from Pre-trained Language Models.} Pre-trained language models~\cite{devlin2019bert,liu2019roberta,lan2019albert} have been demonstrated impressive performance across natural language processing tasks. The implicitly stored knowledge during pre-training can benefit the model on downstream tasks and there have been several papers on evaluating the embedded knowledge in pre-trained language models~\cite{petroni2019language,roberts2020much,talmor2020olmpics,zhou2020evaluating}. This property has been used to solve text classification, natural language inference, relational classification tasks in zero-shot settings by having LMs fill in the blanks of cloze questions~\cite{jiang2020can,schick-schutze-2021-exploiting} or predict the continuation to prompts~\cite{brown2020language,gao2020making}. For example,~\citet{jiang2020can} automatically design several translation rules for extracting 41 different relations and try to ensemble different translation results. We refer the readers to~\citet{liu2021pre} for a more comprehensive survey.

However, these previous work usually focuses on the settings where there are only a fixed set of relations or output classes, and knowledge can be induced by designing a limited amount of hand-crafted or automatically-generated rules, thus these methods cannot be directly applied in commonsense question answering. In this paper, we investigate if this paradigm can also be applied in commonsense question answering and examine ways of adapting natural questions for pre-trained LMs by cloze translation.

\section{Conclusion}

We aim to better utilize the implicitly learned knowledge in pre-trained LMs for commonsense question answering by natural-to-cloze question translation. To this end, we construct a dataset of natural-question pairs and investigate four translation methods. In addition, we demonstrate that different translation methods have distinct characteristics and we propose a consistency optimization objective to combine the strengths of different translations using unlabeled data. We demonstrate the effectiveness of our methods in zero and few-shot settings and show that our methods are complementary to a state-of-the-art knowledge base method. %Analyses also reveal that our methods can be beneficial in high-resource settings when combined with natural questions. 

In the future, we can investigate more cloze translation methods and develop better ways of utilizing the translated questions. Also, so far we only perform experiments on English datasets and our syntactic-based method is specifically designed for English questions, thus it can be interesting to see if our methods can generalize to other languages.

\section*{Acknowledgements}
We would like to thank the anonymous reviewers for valuable suggestions and Da Yin, Nuan Wen for helpful discussions. 
This work is supported by the Machine Common Sense (MCS) program
under Cooperative Agreement N66001-19-2-4032
with the US Defense Advanced Research Projects
Agency (DARPA). The views and the conclusions
of this paper are those of the authors and do not
reflect the official policy or position of DARPA.
%\zd{funding?}

% Use \bibliography{yourbibfile} instead or the References section will not appear in your paper
\bibliography{aaai22}

\begin{thebibliography}{43}
\providecommand{\natexlab}[1]{#1}

\bibitem[{Bauer, Wang, and Bansal(2018)}]{bauer2018commonsense}
Bauer, L.; Wang, Y.; and Bansal, M. 2018.
\newblock Commonsense for Generative Multi-Hop Question Answering Tasks.
\newblock In \emph{Proceedings of the Conference on Empirical Methods in
  Natural Language Processing}.

\bibitem[{Bisk et~al.(2020)Bisk, Zellers, Gao, Choi et~al.}]{bisk2020piqa}
Bisk, Y.; Zellers, R.; Gao, J.; Choi, Y.; et~al. 2020.
\newblock Piqa: Reasoning about physical commonsense in natural language.
\newblock In \emph{Proceedings of the AAAI Conference on Artificial
  Intelligence}, 7432--7439.

\bibitem[{Bosselut, Le~Bras, and Choi(2019)}]{bosselut2019dynamic}
Bosselut, A.; Le~Bras, R.; and Choi, Y. 2019.
\newblock Dynamic knowledge graph construction for zero-shot commonsense
  question answering.
\newblock \emph{arXiv preprint}.

\bibitem[{Bosselut et~al.(2019)Bosselut, Rashkin, Sap, Malaviya, Celikyilmaz,
  and Choi}]{bosselut2019comet}
Bosselut, A.; Rashkin, H.; Sap, M.; Malaviya, C.; Celikyilmaz, A.; and Choi, Y.
  2019.
\newblock {COMET}: Commonsense Transformers for Automatic Knowledge Graph
  Construction.
\newblock In \emph{Proceedings of the Annual Meeting of the Association for
  Computational Linguistics}.

\bibitem[{Brown et~al.(2020)Brown, Mann, Ryder, Subbiah, Kaplan, Dhariwal,
  Neelakantan, Shyam, Sastry, Askell et~al.}]{brown2020language}
Brown, T.~B.; Mann, B.; Ryder, N.; Subbiah, M.; Kaplan, J.; Dhariwal, P.;
  Neelakantan, A.; Shyam, P.; Sastry, G.; Askell, A.; et~al. 2020.
\newblock Language models are few-shot learners.
\newblock \emph{arXiv preprint}.

\bibitem[{Devlin et~al.(2019)Devlin, Chang, Lee, and
  Toutanova}]{devlin2019bert}
Devlin, J.; Chang, M.-W.; Lee, K.; and Toutanova, K. 2019.
\newblock {BERT}: Pre-training of Deep Bidirectional Transformers for Language
  Understanding.
\newblock In \emph{Proceedings of the Conference of the North American Chapter
  of the Association for Computational Linguistics}.

\bibitem[{Gao, Fisch, and Chen(2021)}]{gao2020making}
Gao, T.; Fisch, A.; and Chen, D. 2021.
\newblock Making Pre-trained Language Models Better Few-shot Learners.
\newblock In \emph{Proceedings of the Annual Meeting of the Association for
  Computational Linguistics}.

\bibitem[{Heilman and Smith(2010)}]{heilman2010good}
Heilman, M.; and Smith, N.~A. 2010.
\newblock Good question! statistical ranking for question generation.
\newblock In \emph{Proceedings of the Conference of the North American Chapter
  of the Association for Computational Linguistics}.

\bibitem[{Jiang et~al.(2020)Jiang, Xu, Araki, and Neubig}]{jiang2020can}
Jiang, Z.; Xu, F.~F.; Araki, J.; and Neubig, G. 2020.
\newblock How can we know what language models know?
\newblock \emph{Transactions of the Association for Computational Linguistics}.

\bibitem[{Joshi, Peters, and Hopkins(2018)}]{Joshi2018ExtendingAP}
Joshi, V.; Peters, M.; and Hopkins, M. 2018.
\newblock Extending a Parser to Distant Domains Using a Few Dozen Partially
  Annotated Examples.
\newblock In \emph{Proceedings of the Annual Meeting of the Association for
  Computational Linguistics}.

\bibitem[{Lample et~al.(2018{\natexlab{a}})Lample, Conneau, Denoyer, and
  Ranzato}]{lample2018unsupervised}
Lample, G.; Conneau, A.; Denoyer, L.; and Ranzato, M. 2018{\natexlab{a}}.
\newblock Unsupervised Machine Translation Using Monolingual Corpora Only.
\newblock In \emph{Proceedings of the International Conference on Learning
  Representations}.

\bibitem[{Lample et~al.(2018{\natexlab{b}})Lample, Ott, Conneau, Denoyer, and
  Ranzato}]{lample2018phrase}
Lample, G.; Ott, M.; Conneau, A.; Denoyer, L.; and Ranzato, M.
  2018{\natexlab{b}}.
\newblock Phrase-Based \& Neural Unsupervised Machine Translation.
\newblock In \emph{Proceedings of the Conference on Empirical Methods in
  Natural Language Processing}.

\bibitem[{Lan et~al.(2019)Lan, Chen, Goodman, Gimpel, Sharma, and
  Soricut}]{lan2019albert}
Lan, Z.; Chen, M.; Goodman, S.; Gimpel, K.; Sharma, P.; and Soricut, R. 2019.
\newblock {ALBERT}: A Lite {BERT} for Self-supervised Learning of Language
  Representations.
\newblock In \emph{Proceedings of the International Conference on Learning
  Representations}.

\bibitem[{Lewis et~al.(2020)Lewis, Liu, Goyal, Ghazvininejad, Mohamed, Levy,
  Stoyanov, and Zettlemoyer}]{lewis2020bart}
Lewis, M.; Liu, Y.; Goyal, N.; Ghazvininejad, M.; Mohamed, A.; Levy, O.;
  Stoyanov, V.; and Zettlemoyer, L. 2020.
\newblock {BART}: Denoising Sequence-to-Sequence Pre-training for Natural
  Language Generation, Translation, and Comprehension.
\newblock In \emph{Proceedings of the Annual Meeting of the Association for
  Computational Linguistics}.

\bibitem[{Lewis, Denoyer, and Riedel(2019)}]{lewis2019unsupervised}
Lewis, P.; Denoyer, L.; and Riedel, S. 2019.
\newblock Unsupervised Question Answering by Cloze Translation.
\newblock In \emph{Proceedings of the Annual Meeting of the Association for
  Computational Linguistics}.

\bibitem[{Lin et~al.(2019)Lin, Chen, Chen, and Ren}]{lin2019kagnet}
Lin, B.~Y.; Chen, X.; Chen, J.; and Ren, X. 2019.
\newblock {KagNet}: Knowledge-Aware Graph Networks for Commonsense Reasoning.
\newblock In \emph{Proceedings of the Conference on Empirical Methods in
  Natural Language Processing}.

\bibitem[{Lin et~al.(2020)Lin, Zhou, Shen, Zhou, Bhagavatula, Choi, and
  Ren}]{lin2020commongen}
Lin, B.~Y.; Zhou, W.; Shen, M.; Zhou, P.; Bhagavatula, C.; Choi, Y.; and Ren,
  X. 2020.
\newblock CommonGen: A Constrained Text Generation Challenge for Generative
  Commonsense Reasoning.
\newblock In \emph{Proceedings of the 2020 Conference on Empirical Methods in
  Natural Language Processing: Findings}, 1823--1840.

\bibitem[{Liu et~al.(2021)Liu, Yuan, Fu, Jiang, Hayashi, and
  Neubig}]{liu2021pre}
Liu, P.; Yuan, W.; Fu, J.; Jiang, Z.; Hayashi, H.; and Neubig, G. 2021.
\newblock Pre-train, prompt, and predict: A systematic survey of prompting
  methods in natural language processing.
\newblock \emph{arXiv preprint arXiv:2107.13586}.

\bibitem[{Liu et~al.(2019)Liu, Ott, Goyal, Du, Joshi, Chen, Levy, Lewis,
  Zettlemoyer, and Stoyanov}]{liu2019roberta}
Liu, Y.; Ott, M.; Goyal, N.; Du, J.; Joshi, M.; Chen, D.; Levy, O.; Lewis, M.;
  Zettlemoyer, L.; and Stoyanov, V. 2019.
\newblock Ro{BERT}a: A robustly optimized bert pretraining approach.
\newblock \emph{arXiv preprint}.

\bibitem[{Lv et~al.(2020)Lv, Guo, Xu, Tang, Duan, Gong, Shou, Jiang, Cao, and
  Hu}]{lv2020graph}
Lv, S.; Guo, D.; Xu, J.; Tang, D.; Duan, N.; Gong, M.; Shou, L.; Jiang, D.;
  Cao, G.; and Hu, S. 2020.
\newblock Graph-based reasoning over heterogeneous external knowledge for
  commonsense question answering.
\newblock In \emph{Proceedings of the AAAI Conference on Artificial
  Intelligence}.

\bibitem[{Ma et~al.(2020)Ma, Ilievski, Francis, Bisk, Nyberg, and
  Oltramari}]{ma2020knowledge}
Ma, K.; Ilievski, F.; Francis, J.; Bisk, Y.; Nyberg, E.; and Oltramari, A.
  2020.
\newblock Knowledge-driven Self-supervision for Zero-shot Commonsense Question
  Answering.
\newblock In \emph{Proceedings of the AAAI Conference on Artificial
  Intelligence}.

\bibitem[{Mihaylov et~al.(2018)Mihaylov, Clark, Khot, and
  Sabharwal}]{Mihaylov2018CanAS}
Mihaylov, T.; Clark, P.; Khot, T.; and Sabharwal, A. 2018.
\newblock Can a Suit of Armor Conduct Electricity? A New Dataset for Open Book
  Question Answering.
\newblock In \emph{Proceedings of the Conference on Empirical Methods in
  Natural Language Processing}.

\bibitem[{Omelianchuk et~al.(2020)Omelianchuk, Atrasevych, Chernodub, and
  Skurzhanskyi}]{omelianchuk-etal-2020-gector}
Omelianchuk, K.; Atrasevych, V.; Chernodub, A.; and Skurzhanskyi, O. 2020.
\newblock {GECT}o{R} {--} Grammatical Error Correction: Tag, Not Rewrite.
\newblock In \emph{Proceedings of the Workshop on Innovative Use of NLP for
  Building Educational Applications}.

\bibitem[{Petroni et~al.(2019)Petroni, Rockt{\"a}schel, Riedel, Lewis, Bakhtin,
  Wu, and Miller}]{petroni2019language}
Petroni, F.; Rockt{\"a}schel, T.; Riedel, S.; Lewis, P.; Bakhtin, A.; Wu, Y.;
  and Miller, A. 2019.
\newblock Language Models as Knowledge Bases?
\newblock In \emph{Proceedings of the Conference on Empirical Methods in
  Natural Language Processing}.

\bibitem[{Roberts, Raffel, and Shazeer(2020)}]{roberts2020much}
Roberts, A.; Raffel, C.; and Shazeer, N. 2020.
\newblock How Much Knowledge Can You Pack into the Parameters of a Language
  Model?
\newblock In \emph{Proceedings of the Conference on Empirical Methods in
  Natural Language Processing}.

\bibitem[{Sakaguchi et~al.(2020)Sakaguchi, Le~Bras, Bhagavatula, and
  Choi}]{sakaguchi2020winogrande}
Sakaguchi, K.; Le~Bras, R.; Bhagavatula, C.; and Choi, Y. 2020.
\newblock Winogrande: An adversarial winograd schema challenge at scale.
\newblock In \emph{Proceedings of the AAAI Conference on Artificial
  Intelligence}.

\bibitem[{Sap et~al.(2019{\natexlab{a}})Sap, Le~Bras, Allaway, Bhagavatula,
  Lourie, Rashkin, Roof, Smith, and Choi}]{sap2019atomic}
Sap, M.; Le~Bras, R.; Allaway, E.; Bhagavatula, C.; Lourie, N.; Rashkin, H.;
  Roof, B.; Smith, N.~A.; and Choi, Y. 2019{\natexlab{a}}.
\newblock {ATOMIC}: An atlas of machine commonsense for if-then reasoning.
\newblock In \emph{Proceedings of the AAAI Conference on Artificial
  Intelligence}.

\bibitem[{Sap et~al.(2019{\natexlab{b}})Sap, Rashkin, Chen, LeBras, and
  Choi}]{sap2019socialiqa}
Sap, M.; Rashkin, H.; Chen, D.; LeBras, R.; and Choi, Y. 2019{\natexlab{b}}.
\newblock {SocialIQA}: Commonsense Reasoning about Social Interactions.
\newblock In \emph{Proceedings of the Conference on Empirical Methods in
  Natural Language Processing}.

\bibitem[{Schick and Sch{\"u}tze(2021)}]{schick-schutze-2021-exploiting}
Schick, T.; and Sch{\"u}tze, H. 2021.
\newblock Exploiting Cloze-Questions for Few-Shot Text Classification and
  Natural Language Inference.
\newblock In \emph{Proceedings of the Conference of the European Chapter of the
  Association for Computational Linguistics}.

\bibitem[{Shwartz et~al.(2020)Shwartz, West, Le~Bras, Bhagavatula, and
  Choi}]{shwartz2020unsupervised}
Shwartz, V.; West, P.; Le~Bras, R.; Bhagavatula, C.; and Choi, Y. 2020.
\newblock Unsupervised Commonsense Question Answering with Self-Talk.
\newblock In \emph{Proceedings of the Conference on Empirical Methods in
  Natural Language Processing}.

\bibitem[{Speer, Chin, and Havasi(2017)}]{speer2017conceptnet}
Speer, R.; Chin, J.; and Havasi, C. 2017.
\newblock {ConceptNet} 5.5: An open multilingual graph of general knowledge.
\newblock In \emph{Proceedings of the AAAI Conference on Artificial
  Intelligence}.

\bibitem[{Talmor et~al.(2020)Talmor, Elazar, Goldberg, and
  Berant}]{talmor2020olmpics}
Talmor, A.; Elazar, Y.; Goldberg, Y.; and Berant, J. 2020.
\newblock {oLMpics-On What Language Model Pre-training Captures}.
\newblock \emph{Transactions of the Association for Computational Linguistics}.

\bibitem[{Talmor et~al.(2019)Talmor, Herzig, Lourie, and
  Berant}]{talmor2019commonsenseqa}
Talmor, A.; Herzig, J.; Lourie, N.; and Berant, J. 2019.
\newblock CommonsenseQA: A Question Answering Challenge Targeting Commonsense
  Knowledge.
\newblock In \emph{Proceedings of the Conference of the North American Chapter
  of the Association for Computational Linguistics}.

\bibitem[{Vaswani et~al.(2017)Vaswani, Shazeer, Parmar, Uszkoreit, Jones,
  Gomez, Kaiser, and Polosukhin}]{vaswani2017attention}
Vaswani, A.; Shazeer, N.; Parmar, N.; Uszkoreit, J.; Jones, L.; Gomez, A.~N.;
  Kaiser, {\L}.; and Polosukhin, I. 2017.
\newblock Attention is All you Need.
\newblock In \emph{Proceedings of the Advances in Neural Information Processing
  Systems}.

\bibitem[{Vrande{\v{c}}i{\'c} and Kr{\"o}tzsch(2014)}]{vrandevcic2014wikidata}
Vrande{\v{c}}i{\'c}, D.; and Kr{\"o}tzsch, M. 2014.
\newblock Wikidata: a free collaborative knowledgebase.
\newblock \emph{Communications of the ACM}.

\bibitem[{Wang et~al.(2020)Wang, Peng, Ilievski, Szekely, and
  Ren}]{wang2020connecting}
Wang, P.; Peng, N.; Ilievski, F.; Szekely, P.; and Ren, X. 2020.
\newblock Connecting the Dots: A Knowledgeable Path Generator for Commonsense
  Question Answering.
\newblock In \emph{Proceedings of the Conference on Empirical Methods in
  Natural Language Processing: Findings}.

\bibitem[{Wang, Ruder, and Neubig(2021)}]{wang21naacl}
Wang, X.; Ruder, S.; and Neubig, G. 2021.
\newblock Multi-view Subword Regularization.
\newblock In \emph{Proceedings of the Conference of the North American Chapter
  of the Association for Computational Linguistics}.

\bibitem[{Xu et~al.(2021)Xu, Zhu, Xu, Liu, Zeng, and Huang}]{xu2020fusing}
Xu, Y.; Zhu, C.; Xu, R.; Liu, Y.; Zeng, M.; and Huang, X. 2021.
\newblock Fusing Context Into Knowledge Graph for Commonsense Reasoning.
\newblock In \emph{Proceedings of the Annual Meeting of the Association for
  Computational Linguistics}.

\bibitem[{Ye et~al.(2019)Ye, Chen, Wang, and Ling}]{ye2019align}
Ye, Z.-X.; Chen, Q.; Wang, W.; and Ling, Z.-H. 2019.
\newblock Align, mask and select: A simple method for incorporating commonsense
  knowledge into language representation models.
\newblock \emph{arXiv preprint}.

\bibitem[{Zellers et~al.(2019)Zellers, Bisk, Farhadi, and
  Choi}]{zellers2019recognition}
Zellers, R.; Bisk, Y.; Farhadi, A.; and Choi, Y. 2019.
\newblock From recognition to cognition: Visual commonsense reasoning.
\newblock In \emph{Proceedings of the IEEE Conference on Computer Vision and
  Pattern Recognition}.

\bibitem[{Zellers et~al.(2018)Zellers, Bisk, Schwartz, and
  Choi}]{zellers2018swag}
Zellers, R.; Bisk, Y.; Schwartz, R.; and Choi, Y. 2018.
\newblock {SWAG}: A Large-Scale Adversarial Dataset for Grounded Commonsense
  Inference.
\newblock In \emph{Proceedings of the Conference on Empirical Methods in
  Natural Language Processing}.

\bibitem[{Zhou et~al.(2019)Zhou, Khashabi, Ning, and Roth}]{zhou2019going}
Zhou, B.; Khashabi, D.; Ning, Q.; and Roth, D. 2019.
\newblock “Going on a vacation” takes longer than “Going for a walk”: A
  Study of Temporal Commonsense Understanding.
\newblock In \emph{Proceedings of the 2019 Conference on Empirical Methods in
  Natural Language Processing and the 9th International Joint Conference on
  Natural Language Processing (EMNLP-IJCNLP)}, 3354--3360.

\bibitem[{Zhou et~al.(2020)Zhou, Zhang, Cui, and Huang}]{zhou2020evaluating}
Zhou, X.; Zhang, Y.; Cui, L.; and Huang, D. 2020.
\newblock Evaluating commonsense in pre-trained language models.
\newblock In \emph{Proceedings of the AAAI Conference on Artificial
  Intelligence}.

\end{thebibliography}

\end{document}